\documentclass{article}
\usepackage{graphicx}
\usepackage{amsmath,amssymb} % define this before the line numbering.
\usepackage{color}
\usepackage{url}
\usepackage{comment}
\usepackage{subcaption}

\usepackage{algorithm,algorithmic}
\usepackage[pagebackref=true,breaklinks=true,letterpaper=true,colorlinks,bookmarks=false]{hyperref}

\usepackage[final]{corl_2018}
	 
\title{Reinforcement Learning of Active Vision for  Manipulating Objects under Occlusions}

\author{
  Ricson Cheng, Arpit Agarwal, Katerina Fragkiadaki\\
  Carnegie Mellon University \\
  United States\\
  \texttt{\{ricsonc,arpita1\}@andrew.cmu.edu, katef@cs.cmu.edu } \\
}

\newcommand{\QQ}{\mathrm{Q}}
\newcommand{\st}{\mathrm{s}}
\newcommand{\obs}{\mathrm{o}}

\newcommand{\act}{\mathrm{a}}
\newcommand{\actg}{\mathrm{a_g}}
\newcommand{\actc}{\mathrm{a_c}}
\newcommand{\go}{\mathrm{g}}

\begin{document}

\maketitle
\begin{abstract}
We consider artificial agents that learn to jointly control  their gripper and  camera in order to reinforcement learn manipulation policies in the presence of occlusions from  distractor objects. 
Distractors often occlude the object of interest and cause it to disappear from the field of view.  
We propose  hand/eye controllers that learn to move the camera to keep the object within the field of view and visible, in coordination to manipulating it to achieve the desired goal, e.g.,  pushing it to a target location.  We incorporate structural biases of object-centric attention within our actor-critic architectures,  which our experiments suggest to be a key for good performance. 
Our results further highlight the importance of curriculum with regards to environment difficulty. The  resulting active vision / manipulation policies outperform static camera setups for a variety of cluttered environments. 
\end{abstract}

%that feedback to each other via state abstractions: the observer considers the current image and the chosen manipulation action and selects the next camera motion. This in turn generates the next visual image, which enters an object detector that predicts the object location, which is the input for the manipulator actor network. 

\begin{comment}
   We consider reinforcement learning manipulation policies in the presence of occlusions from environmental distractors. We show how such ubiquitous in real life setup poses challenges to assumptions of known and easily obtained state of current methods. We propose agent archotectures that learn to jointly control their manipulator as well as their camera in order to handle occlusions and learn to estimate the 3d object state from actively framed RGB images. We compare against static camera agents of current works, as well as non modular actor-observer architectures.
\end{comment}

\section{Introduction}

%currently state estimation either known or trivial in each frame-no memory

%humans seek best viewpoints and coordinate vision and action. 
We consider artificial agents that learn to perform manipulation tasks in cluttered environments. In this case, state estimation, namely the prediction of 3D object locations and poses, is particularly challenging due to frequent occlusions of the target object from distractors. %We take inspiration from the way humans handle the difficult state estimation problem. 
We, humans, move our head to find the best viewpoint for the actions we wish to  carry out. We turn our heads towards the direction we expect  the object to move, and actively select viewpoints to facilitate perception during manipulation. 
%be found as a result of their actions. In order words, humans actively move their eyes to facilitate action in the presence of occlusions. 
Yet, reinforcement learning has thus far almost exclusively  considered static cameras. If the agent was given the opportunity to (learn to)  move its camera, potentially better manipulation policies would be found, exploiting the easier state estimation thanks to dis-occlusions of the object of interest by active vision policies \cite{soat}. 

%overview and surprising results
%We claim that state estimation needs to be an integral part of robotic learning, and 

We explore  actor-critic  architectures for %action and active perception
%architectures for 
%reinforcement learning that couple perception and action, by learning 
hand/eye control in cluttered scenes. We specifically consider pushing an object to target locations in environments with distractors; the distractors often cause the object of interest to be  occluded. The proposed architectures learn camera control policies that facilitate state estimation by  learning to increase visibility of the main object the agent interacts with, in coordination to the manipulation actions of the agent.  
%guided by rewards regarding action completion. 
%We propose modular neural actor-critic  architectures for action and active perception and show importance design choices for their stable training.  
%We train hand/eye control policies for pushing an object around in environments with distractors, which frequently cause the object to be  occluded. 
%\todo{rewrite this}
Our experiments suggest two important and surprising facts:
%\begin{enumerate}
 %\item 
 First, training with vanilla CNN architectures failed to learn the task of object pushing in the presence of even simple distractors. Incorporating a trained  object detector module in the actor-critic architecture
 %, as well as exposing  only the relevant  object-centered part of the environment to the gripper policy, r
 results in effective learning of the manipulation task under occlusions. %Moreover, exposing only the relevant 
    %\item State abstraction. Only part of the 
    Second, state-of-the-art reinforcement   learning methods  failed to learn the task of object pushing in the presence of even simple distractors, that occasionally occlude the object. Initializing the actor-critic network weights from policies learned in environments  without distractors, quickly and effectively raises success rate in the difficult environment. This highlights the importance of curriculum learning regarding environment difficulty.
   
%\end{enumerate}
Our resulting hand-eye control policies outperform  manipulation policies learned with a static camera. 
%This highlights the importance of curriculum learning regarding environment difficulty. 
%Second,  learning to control the camera viewpoint solely guided by the extrinsic task reward provides an additional performance boost  in the presence of distractors. %, that combine object detectors, camera motion selection, and gripper motion  and show that state abstraction, i.e., not providing the full image as input to the manipulator actor, but rather, compute state and provide only the state input, permits effective learning of perception and action. 
%carried out by an object detector. 
%methods, and  
%The fielf od reinforcement learning for object manipulation often starts by assuming object 
%summary
In summary our contributions are as follows:
\begin{itemize}
    \item We introduce the problem of  learning   manipulation policies under occlusions,  and propose agents that can  control both hand and eye movement, to coordinate active perception and action, in environments with various types of distractors. 
    \item We present modular actor-critic network architectures for  action and perception in which only part of the state is exposed to the gripper controller, and where object detector modules are used to localize the object in the selected camera viewpoints.  
    %the manipulator  provides as input the action to the camera actor to adjust in the next time step, and the camera actor provides the raw RGB input to object detectors to detect objects in the scene and feed their locations to the manipulator. 
    The proposed modular  architectures outperforms non-modular alternatives. 
    %that better and faster learns as oppposed to joint actor-critic that predicts  
   % \item We present extensive experimental results regarding the importance of active perception, curriculum learning and object-centric structural bias  for reinforcement learning. %, in convincing experimental setups.
\end{itemize}

Our code will be publicly available at  \url{https://github.com/ricsonc/ActiveVisionManipulation}. %upon acceptance, and it is also included in the supplementary material.

\section{Related work}
\label{sec:related}

% \todo{we need to put this somewhere}
% %State estimation, namely the prediction of 3d object locations and poses, is often neglected in current robotic learning methods. 
% In reinforcement learning for manipulation, objects 3D centroids are often assumed known \cite{andrychowicz2017hindsight}, or trivially obtained from RGB input \cite{DBLP:journals/corr/abs-1710-01813,DBLP:journals/corr/abs-1710-06542}. This assumes there is no severe occlusion between the manipulated objects, the gripper or other distractors present in the scene. Often times, multiview environments and/or AR tags are used to track  objects locations and poses over time. \cite{Hashimoto_touchme:an}. 

\paragraph{State estimation and reinforcement learning}
%\todo{arpit should carefully fill in the citations here!}
Many Reinforcement Learning (RL) control methods manually design the state representation, that is often comprised of object and gripper 3D locations and poses, velocities, etc.  which they  assume given \cite{andrychowicz2017hindsight}, or easily obtainable with an object detector \cite{DBLP:journals/corr/abs-1710-01813,DBLP:journals/corr/abs-1710-06542}.  This often assumes there is no severe occlusion between the manipulated objects, the gripper or other distractors present in the scene. %Often times, multiview environments and/or AR tags are used to track  objects locations and poses over time. \cite{Hashimoto_touchme:an}.  
Tobin et al. \cite{DBLP:journals/corr/TobinFRSZA17} use synthetic data augmentation to learn detectors in cluttered environments. However, objects are almost never fully occluded in their setup, so is it not necessary for the camera to be active. Ebert et al.  \cite{DBLP:journals/corr/abs-1710-05268} learns a frame prediction model that can handle occlusions by using unoccluded past views to move the objects from. Instead, we take the complementary approach of keeping the object of interest in view using an active camera.

Other works attempt control directly from RGB images \cite{DBLP:journals/corr/abs-1710-06542}. It has been shown that such frame-centric representations do not generalize under environmental variations between training and test conditions \cite{DBLP:journals/corr/KanskySMELLDSPG17}. Our manipulation policy takes as input an RGB image, but also separately estimates the pose of objects in the scene using an object detector. We control the camera motion to facilitate the work of the object detector module, under occlusions from the environment.

%new section

%

\paragraph{Active vision}
An active vision system is one that can manipulate the viewpoint of the camera(s) in order to investigate the environment and get better information from it \cite{ap,aor}. Active visual recognition attracted attention as early as 1980's. The work of \cite{activealoimonos} suggested many problems such as shape from contour and structure from motion, while ambiguous for a passive observer, they can easily be solved for an active one \cite{1238372,blur}. 
%Researchers attempted to select zoom in/zoom out settings for object tracking\cite{1238372}, minimize motion blur \cite{blur}, enhancing depth perception of an object by focusing two cameras on the same object or moving the cameras \cite{activealoimonos}. 
Active control of the camera view point also helps in focusing computational resources on the relevant element of the scene  \cite{doi:10.1167/11.5.5}. The importance of active data collection for sensorimotor coordination has been highlighted in the famous experiment of Held and Hein  in 1963 \cite{kitten} involving two kittens engaged soon after birth to move and see in a carousel apparatus. The active kitten that could move freely was able to develop healthy visual perception, but the passively moved around kitten suffered significant deficits. %one was  the rolling cats: though two kitten observed the same series of visual scenes, the kitten that was selecting the scenes learnt while the kitten that was carried around passively obsering was not able to learn.
%\todo{can someone collect active vision references maybe fro our related paper? We need some very old citations, and how these old works are different than current reinforcement learning methiods}
%The birth of Computer Vision in fact started with approaches that were trying to decide where to look for obstacle avoidance as opposed to explain what they are looking at. Such methods were active vision as they were developed having in mind an active robotic agents. 
Despite its  importance, active vision was put aside potentially due to the slow development of autonomous robotics agents, necessary for the realization of its full potential. 
%Many researchers shifted attention to recognition from large static image collections \cite{} and use of big data neural networks. 
Recently, the active vision paradigm has been resurrected by approaches that perform selective attention \cite{where} and selective image glimpse  processing in order to accelerate   object localization  and recognition    in static images \cite{mathe,fer,aolw}.  Calli et al. \cite{Calli8310020}  select camera  views in order to maximize the success rate of object classification. Fewer methods have addressed    camera view selection for a moving observer \cite{dfd,tdvn,DQL}, e.g., for navigation \cite{kavu,gup} or for object 3D reconstruction in synthetic scenes \cite{DBLP:journals/corr/abs-1709-00507}. Radmard et. al \cite{Radmard8304825} explicitly maps out occluded space in order to determine how to move the camera. 
%Van Hoof et al. \cite{van2014probabilistic} demonstrates a method for probabilistic segmentation of a scene, where the motion of the objects or multiple views can be used to aid segmentation. Whereas in van Hoof et al.  segmentation is the end goal, in our work, selecting a good view for the RCNN/perception is  an intermediate goal which helps us achieve the end goal of pushing an object. 
%Calli et al. \cite{Calli8310020}  select camera  views in order to maximize the success rate of object classification. 
%Instead of classification, we attempt to learn a manipulation policy. 
%They use an extremum seeking control algorithm whereas we learn our camera policy using reinforcement learning. 
%Radmard et. al \cite{Radmard8304825} explicitly maps out occluded space in order to determine how to move the camera. 
Our approach learns such camera control guided by a reward of achieving a manipulation task, as opposed to manually designing such policy. Gualtieri and Platt \cite{2018arXiv180606134G} consider a robot that can focus on different regions of a precaptured pointcloud of the entire scene while performing a manipulation task. We do not assume pointcloud input, instead we learn to control the camera directly from RGB input. 
%In fact, adding auxiliary visibility rewards did not help in our framework, as described in the experimental section. 
%Gualtieri and Platt \cite{2018arXiv180606134G} make use of gaze actions in which the robot can focus on different regions of a precaptured pointcloud of the entire scene while performing a manipulation task. We do not assume pointcloud input, instead we learn to control the camera directly from RGB input. 
%object recognition and localization methods in 2d images 
%have re
%image based reinforcement learning methods have resurrected the old idea \cite{} by learning to select camera in synthetic environments: using 3d synthetic object models, they attempt to choose the next best view to improve 3d object reconstruction. 
%Many of the aforementioned approaches  use some type of recurrent network \cite{mathe,fer,aolw,DBLP:journals/corr/MnihHGK14,LSM} that  accumulates information across views or glimpses. 
%The latent state  is then mapped to a 3d voxel occupancy grid \cite{DBLP:journals/corr/abs-1709-00507} or object category label \cite{}. 
%Since we cannot backpropagate through view or glimpse  selection, reinforcement learning methods have been often used to train such view or glimpse selectors.  
To the best of our knowledge, our work is the first (in recent years) to consider \textbf{active perception for action, as opposed to recognition.} 
%While the availability of large standardized datasets much facilitates machine perception on internet images, machine perception for action holds great premise on fine-grained visual understanding thanks to the possibility of actively selecting the camera viewpoint.

\section{Reinforcement learning hand/eye coordination for object manipulation}

We consider artificial agents that can control both their camera and gripper to complete manipulation tasks. We  specifically investigate pushing objects to particular locations in the workspace in the presence of distractors, that may occlude occasionally the object of interest.  Our proposed architectures  can be extended to  other  manipulation tasks. 

We consider a multi-goal Partially Observable Markov Decision Process (POMDP), 
represented by observations  $\obs \in \mathcal{O}$, states $\st \in \mathcal{S}$, goals $\go \in \mathcal{G}$, hand (gripper) actions $\actg \in \mathcal{A^G}$, and eye (camera)  actions $\actc \in \mathcal{A^C}$. Let  $\mathcal{A}: \mathcal{A^C} \times \mathcal{A^G}$ denote the full action space.  
At the beginning of each episode, an  observation-goal pair is sampled from the initial observation distribution $p(\obs_0,\go)$. 
Each goal corresponds to a reward function $r_g : \mathcal{S} \times \mathcal{A} \rightarrow \mathbb{R}$. At each timestep of an episode, the agent gets as input the current observation and goal, and chooses actions according to policy $\pi: \mathcal{O} \times \mathcal{G} \rightarrow \mathcal{A}$ which results in reward  $r_t = r_g(\obs_t,\act_t)$. The objective of the agent is to
maximize the expected discounted future reward. 

\begin{figure}[h]
    \centering
    \begin{minipage}{0.95\textwidth}
        \centering
        \includegraphics[width=\textwidth]{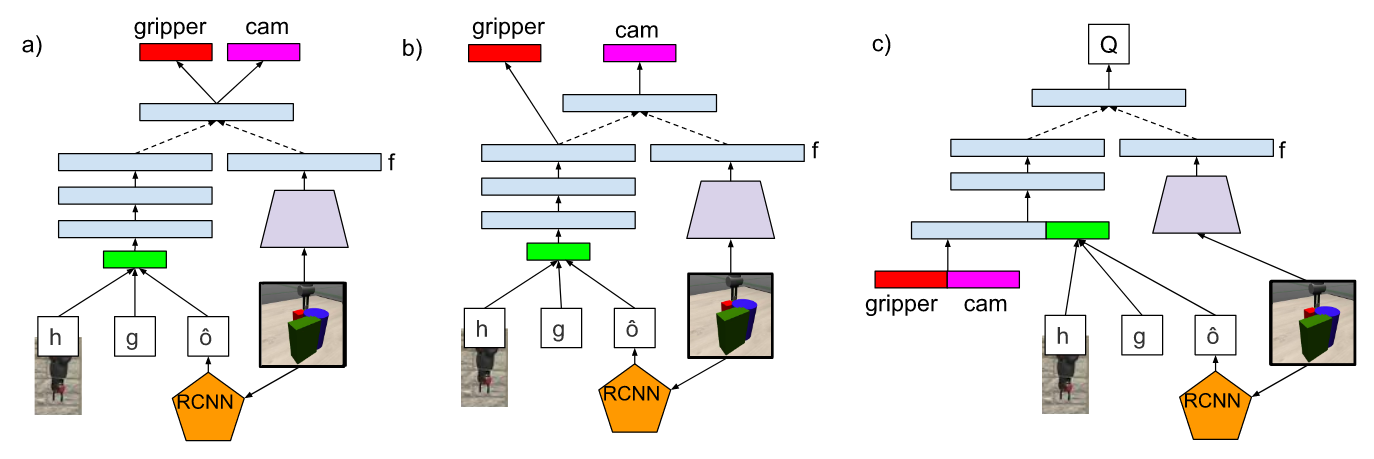}
    \end{minipage}
    \begin{minipage}{0.95\textwidth}
        \centering
        \includegraphics[width=\textwidth]{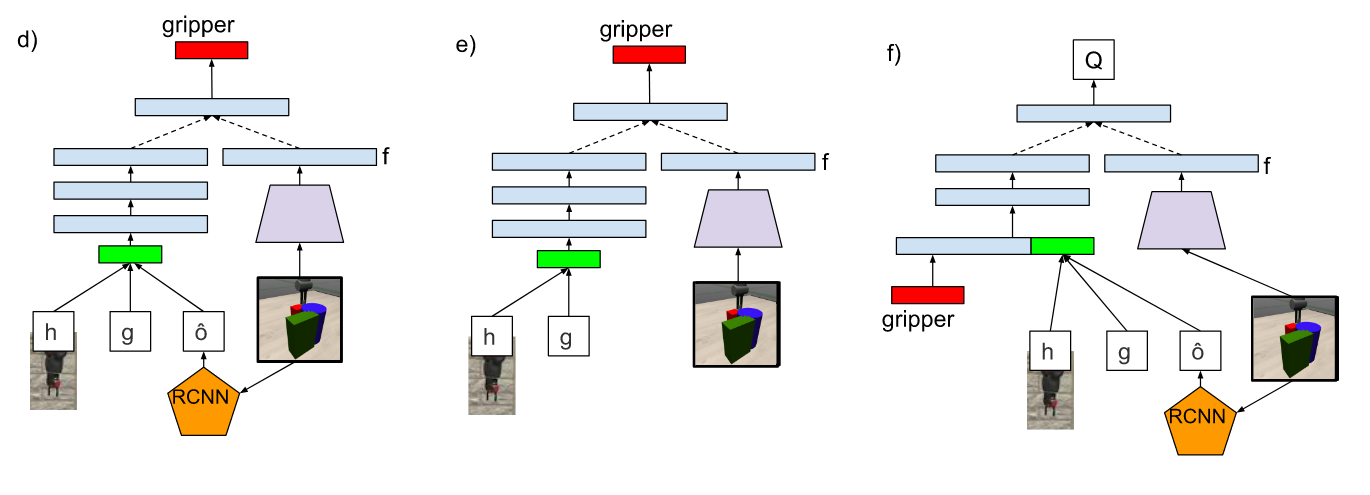}
    \end{minipage}
    \caption{ \textbf{Actors and critic architectures for active} \textit{(top row)} \textbf{and static} \textit{(bottom row)} \textbf{ camera agents}. a) Actor network for {\it cam-active-full}. The same state representation is fed as input to hand and eye actor networks. b) Actor network for {\it cam-active-abstr}. The gripper actor  responsible for pushing the object does not receive direct information from the RGB  input, but only the 3D location of the object. c) Critic network for all active camera agents. d) Actor network for {\it cam-static}, the agent can only control its gripper, not its camera. e) Actor network for {\it cam-static-image}. The pretrained object detector subnetwork is omitted. f) Critic network for {\it cam-static}.  Dotted lines denote elementwise addition. Trapezoids denote convolutional neural sub-networks, and blue rectangles denote fully connected layers. }
    \label{fig:arch}
\end{figure}

Our hand-eye controller architectures are represented by (generalized) modular actor-critic networks. 
We explored two distinct actor architectures, illustrated in Figure \ref{fig:arch} (a), (b), both of which we couple with the same critic architecture, illustrated in  Figure \ref{fig:arch}  (c).  

The critic network takes as input an observation, namely, an RGB frame $I_t$, and encodes it using a what-where decomposition.
Specifically, the input observation is encoded into:
\begin{itemize}
    \item a low-dimensional embedding vector $f_t \in \mathbb{R}^{64}$ that represents the frame appearance. We use a convolutional neural network (CNN) depicted as a purple trapezoid in Figure \ref{fig:arch} (c). 
    \item a 3D centroid for the object of interest $\hat{o}_t \in \mathbb{R}^{3}$.  A visual detector module  \cite{DBLP:journals/corr/RenHG015}  outputs a 2D bounding box for the object of interest.  We recover the 3D location of the object by considering the camera ray which passes through the center  of the detected 2D bounding box, and intersecting it with the horizontal plane on which the object lies, using ground-truth camera intrinsic and extrinsic parameters.
\end{itemize}

 The predicted object location $\hat{o}_t$ is fed alongside the gripper location $h_t$ (assumed known) and the desired goal object location $\go$ to the critic, alongside gripper and camera actions $\actg_t,\actc_t$, and the RGB image embedding $f_t$. Our (state, goal) representation thus reads: 
  \begin{equation}
  (\st_t,\go)=([f_t;\hat{o}_t,h_t],\go),
     \label{eq:fc}
 \end{equation}
   where $\hat{o}_t,h_t,\go$ stand for absolute 3D locations in the workspace. Empirically, we have found  an object-centric representation for the object and gripper locations to lead to faster learning, namely:
 \begin{equation}
  (\st_t,\go)=([f_t;h_t-\hat{o}_t],\go-\hat{o}_t).
     \label{eq:oc}
 \end{equation}
Such object-centric representation  exploits translation invariance of the physical laws, and thus of the resulting policies, and has been used in previous works \cite{andrychowicz2017hindsight}. 
This invariance is mostly true in our environment as long as we do not go beyond the limits of the gripper. We use this object-centric representation in all our experiments.

 \textbf{A moving camera drastically increases the amount of variability exhibited in the input observation sequences}, making it  harder to learn  manipulation policies directly from full frame input. We experimented with actor architectures which do not use the frame appearance embedding as input to the gripper actor policy, rather only the location of the object of interest. 
 Specifically, we explore the following two actor architectures:
 \begin{enumerate}
     \item \textit{Full} state information available to the gripper actor network (Figure \ref{fig:arch} (a)).  
     Frame appearance embedding $f_t$, object location $\hat{o}_t$ and gripper location $h_t$, are provided to both hand and eye actor subnetworks.   
     \item  State \textit{abstraction} for the gripper actor network (Figure \ref{fig:arch} (b)). The image embedding $f_t$ is provided as input only to  the eye control policy, while the object and gripper locations are provided as input to both (hand and eye) controllers. In this way, the gripper actor network uses a  manual yet way less variable state representation,  which may be simpler to learn from. 
 \end{enumerate}
 
Our hand/eye controllers are trained with reinforcement learning and  Hindsight Experience Replay \cite{andrychowicz2017hindsight} (HER). HER introduced the powerful idea that failed executions -- episodes that do not achieve the desired goal -- achieve some certain goal, and it is useful to book-keep such failed experience as successful experience for that alternative goal in the experience buffer. Such multi-goal experience is used  to train a generalized policy, where actor and critic networks take as input both the current state as well as the current goal (as opposed to the state alone). Thanks to the smoothness of actor and critic networks, achieved goals that are nearby the desired goal, are used to learn useful value and actions, instead of being discarded.  
Training is carried out with off-policy deep deterministic policy gradients (DDPG)  \cite{DBLP:journals/corr/LillicrapHPHETS15}. %This allows us to decouple exploration and policy learning.
The agent maintains actor $\pi: \mathcal{S}   \rightarrow \mathcal{A}$ and action-value (critic)  $\QQ : \mathcal{S} \times \mathcal{G}\times \mathcal{A} \rightarrow \mathbb{R}$ function approximators. The actor is learned by taking gradients with loss function $\mathcal{L}_a = - \mathbb{E}_s \QQ(\st,\go,\pi(\st,\go))$ and the critic minimizes TD-error using TD-target  $y_t = r_t + \gamma \QQ(\st_{t+1}, \go, \pi(\st_{t+1},\go))$, where $\gamma$ the reward discount factor. Exploration is carried out by adding $\epsilon$ normal stochastic noise to actions predicted by current policy.  HER has shown to outperform  \textit{single goal} policy learning with deep deterministic policy gradients (DDPG)  \cite{DBLP:journals/corr/LillicrapHPHETS15}, and is currently a state-of-the-art RL method for continuous control. Since our goal representation is desired locations in the workspace the object should reach, HER is particularly well suited for our setup, since we do expect learned policies to change smoothly with respect to goal locations.

\paragraph{Implementation details}

The convolutional network  used in the  actor and critic networks and depicted as purple trapezoid in Figure \ref{fig:arch} takes as input a $64 \times 64 \times 3$ image. The network has 3 layers with 16, 32, and 64 kernels of size $4 \times 4$ with strides 2, 2, and 4. This is followed by a global average pooling layer, then a final linear layer with 64 units is applied. Batch normalization is used after every convolution layer.

All the fully connected layers in the actor and critic networks have 64 units each. Layer normalization and relu activation are used. The action output layer uses tanh activation function. In the critic architecture, the action is concatenated with the object-centric state representation  after one layer.

At test time, we use an RCNN \cite{DBLP:journals/corr/RenHG015} as the object detector with a ResNet-101 backbone. The RCNN detector module is initialized with  weights obtained from training for the MS COCO object detection benchmark \cite{DBLP:journals/corr/LinMBHPRDZ14}, and further finetuned to detect the object of interest in 10000 initial states of our environments. We use amodal 2D bounding boxes as ground-truth. This allows the RCNN to infer the position of an object even under heavy occlusion. We randomly initialized the camera position while collecting the training data. The weights of the object detector module are fixed during reinforcement learning. 
Since it is computationally prohibitive to use the detector module during training, we substitute it with a simulated RCNN. The simulated RCNN detects the object with probability $p = \min(4.6v, 0.1v+0.9)1-\sqrt{\text{occ}}$ where $v\text{occ}$ is the fracproportion of the object mask which is unoccluded. When the object is detected, we add random normal noise $(\mu = 0, \sigma = 1)$ to the bounding box coordinates and return that as the output. We found no improvement from fine-tuning the final policy using the real RCNN detector module, which is thus used only at test time. A much lighter detector convolutional backbone could have been used instead. 

\begin{figure}[h]
    \centering
    \setlength{\tabcolsep}{0pt}
    \begin{tabular}{ccc}
        \includegraphics[trim={1cm 1cm 1cm 1cm},width=0.33\textwidth]{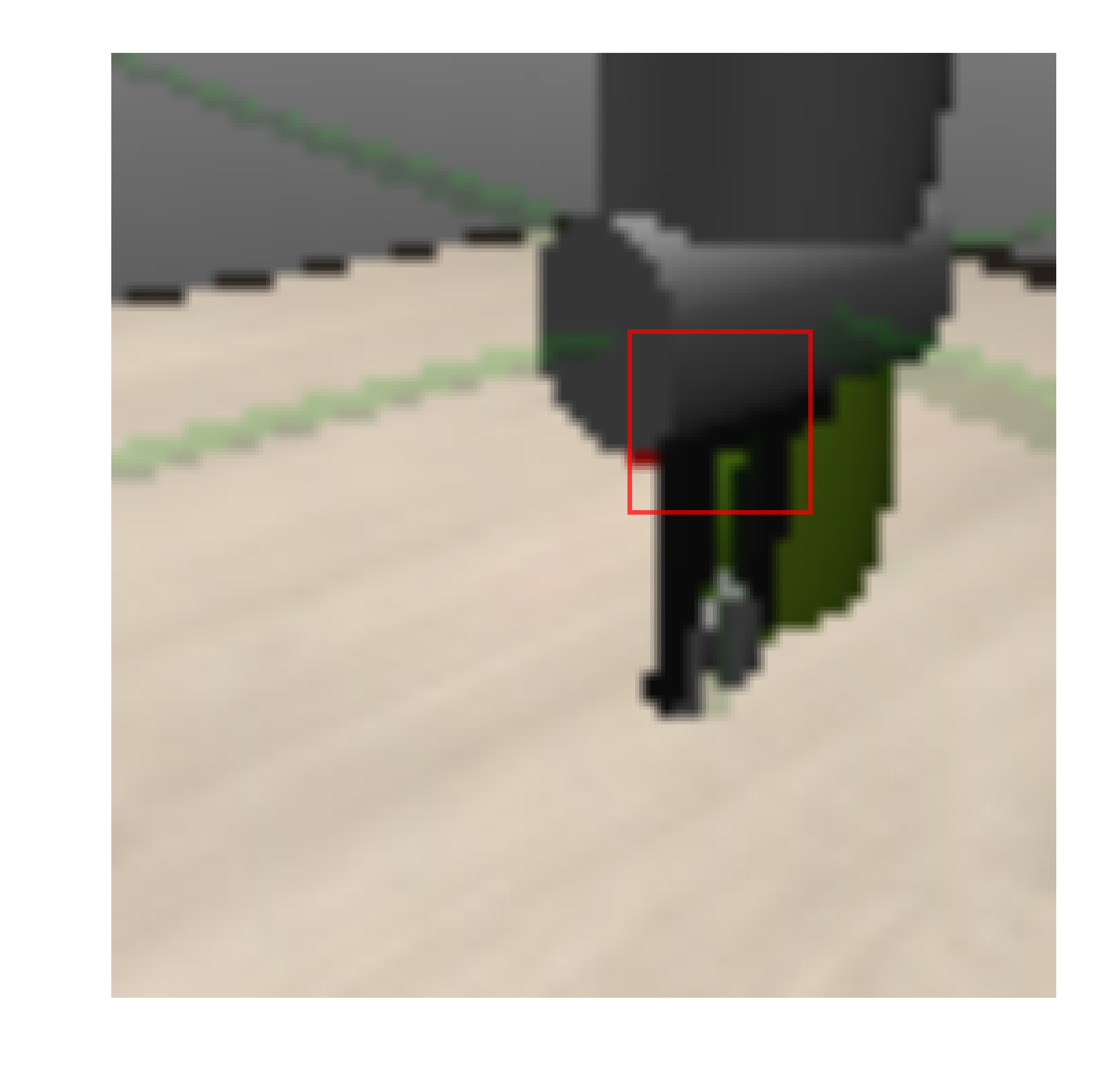} & \includegraphics[trim={1cm 1cm 1cm 1cm},width=0.33\textwidth]{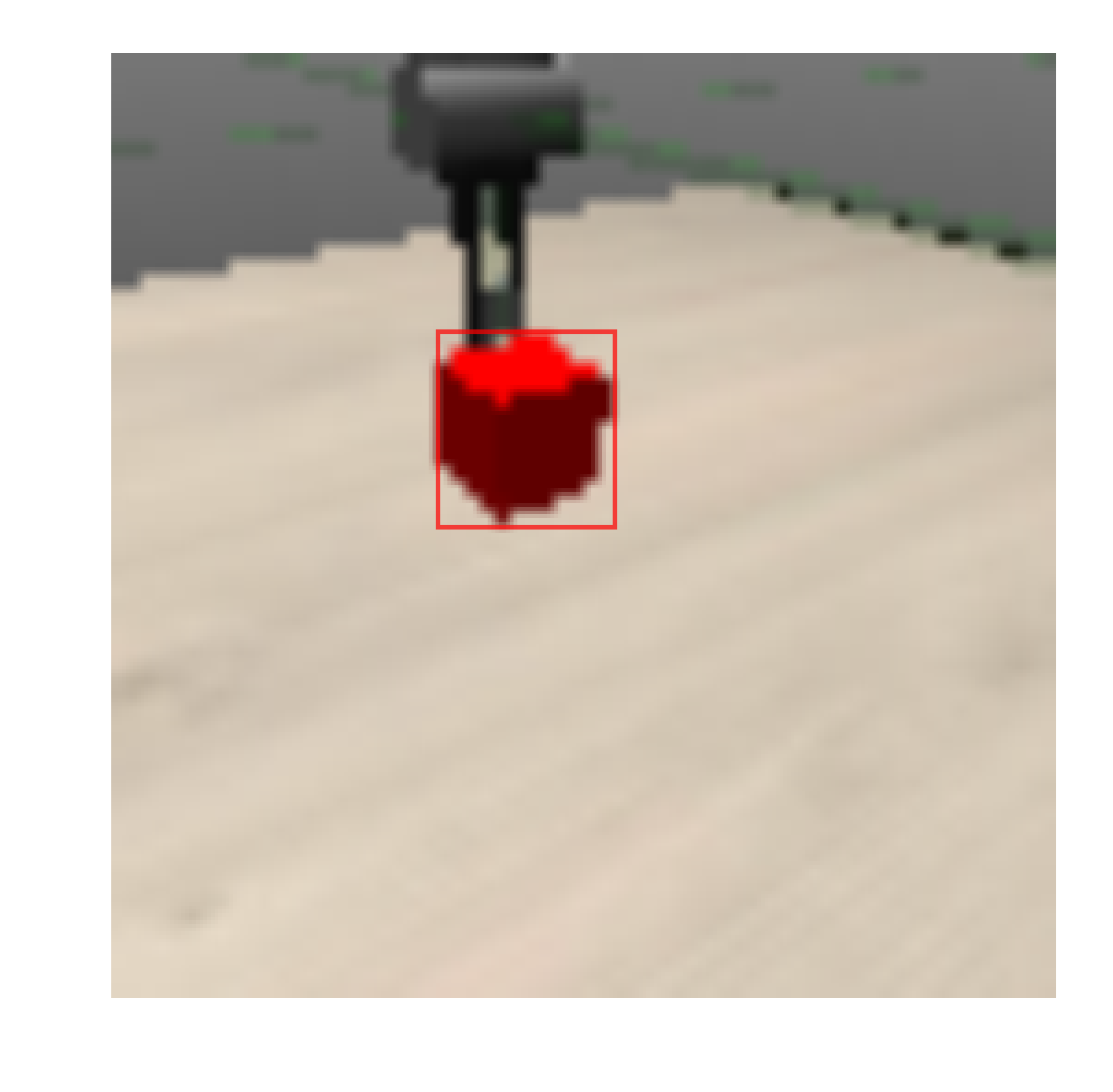} & \includegraphics[trim={1cm 1cm 1cm 1cm},width=0.33\textwidth]{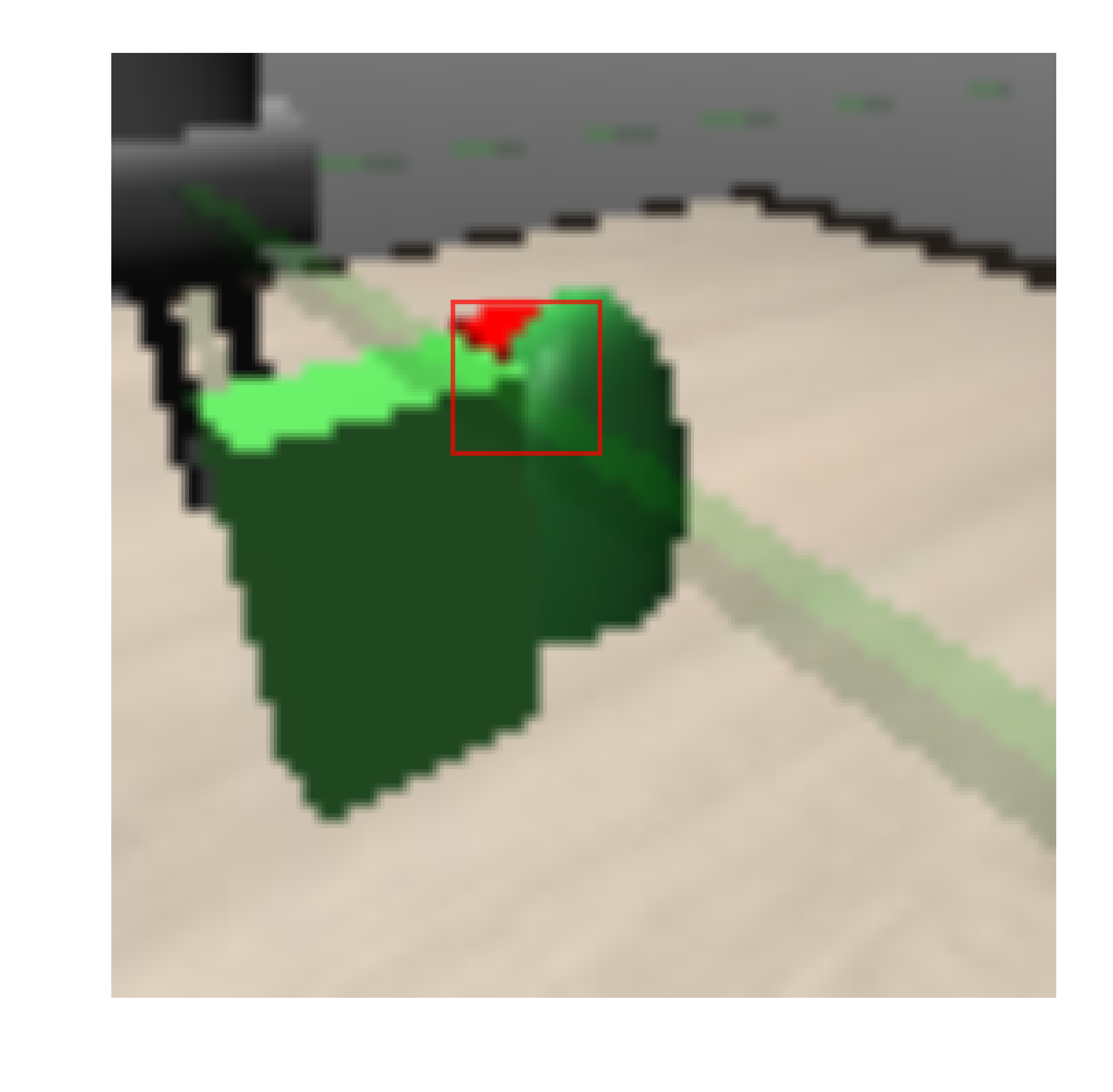} \\
        \includegraphics[trim={1cm 1cm 1cm 1cm},width=0.33\textwidth]{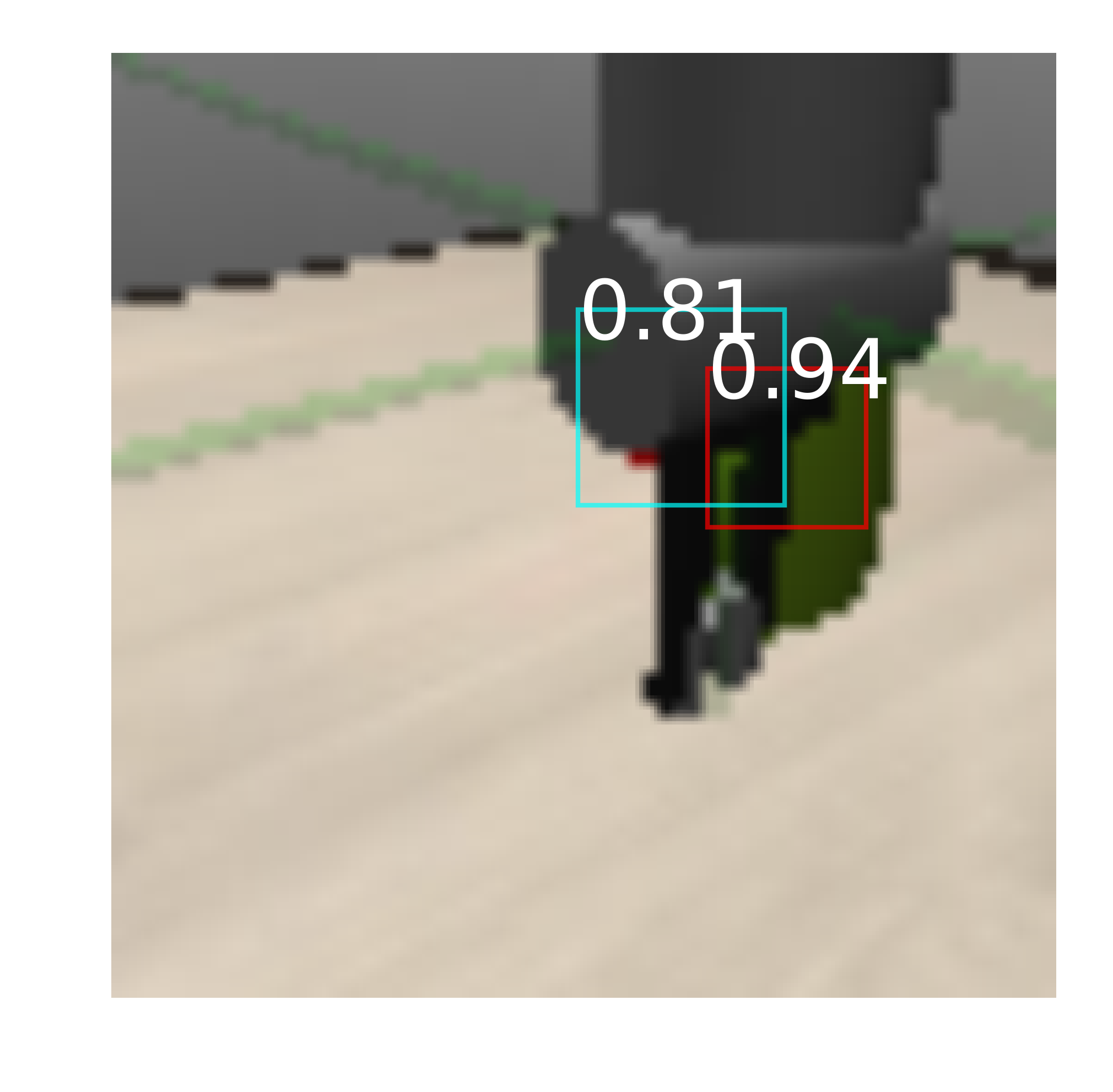} & \includegraphics[trim={1cm 1cm 1cm 1cm},width=0.33\textwidth]{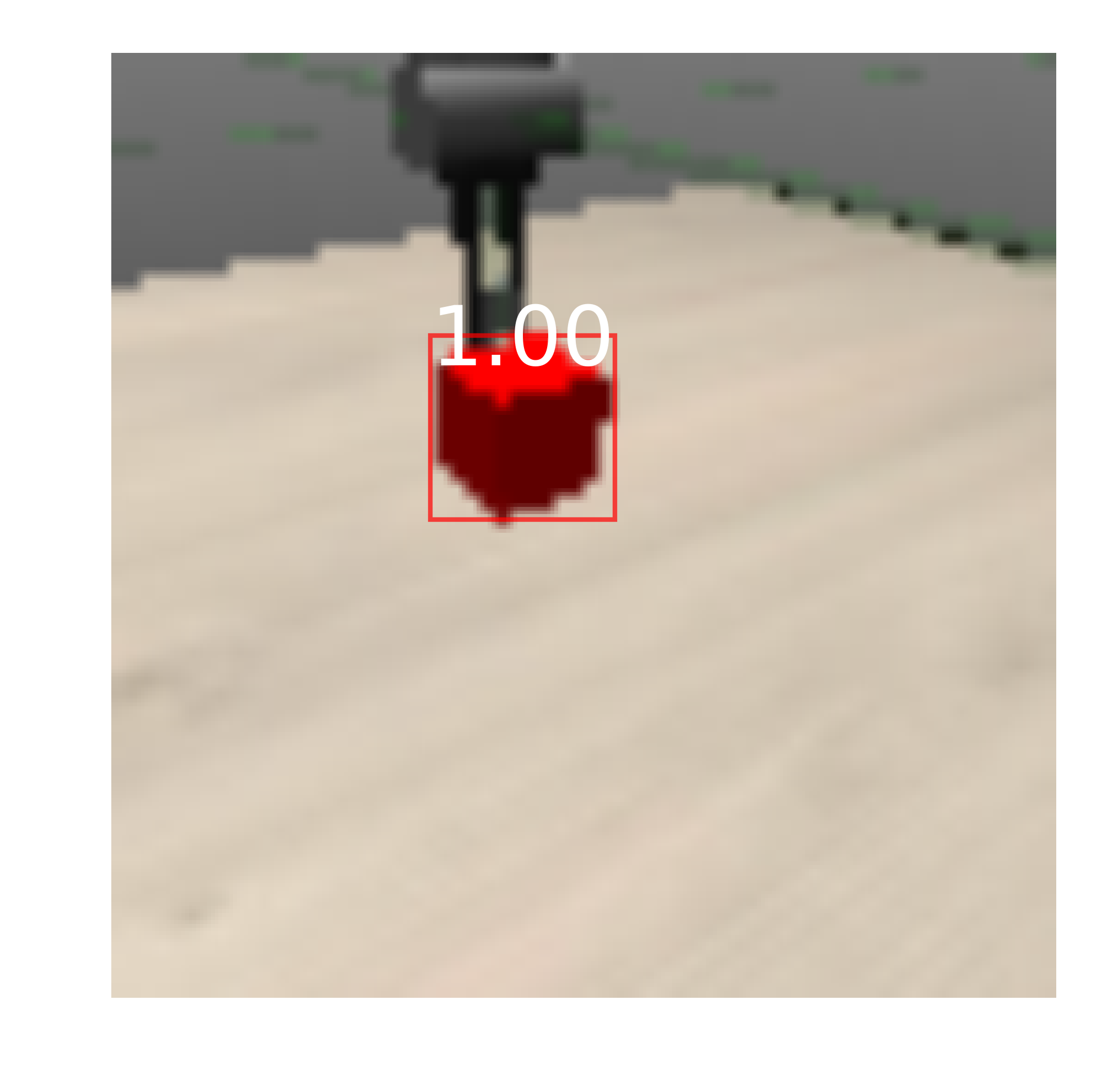} & \includegraphics[trim={1cm 1cm 1cm 1cm},width=0.33\textwidth]{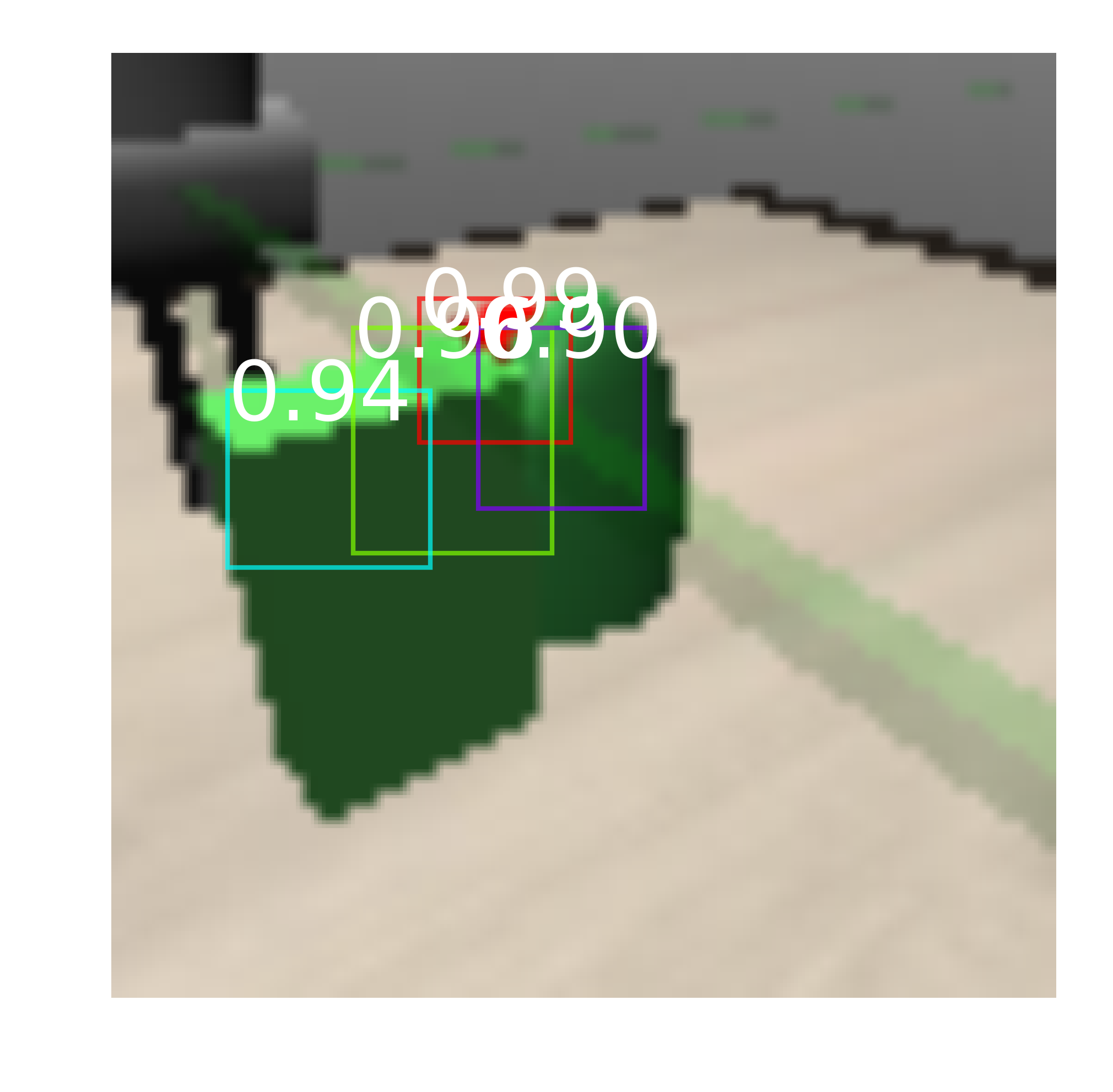} \\
    \end{tabular}
    \caption{ Sample predictions made by the trained RCNN. The RCNN is trained using {\it amodal} boxes. First row: ground truth, Second row: predictions, with confidence scores shown. Note that the RCNN is capable of accurately inferring the amodal boxposition of the object even when it is highcompletely occluded, but not when it is completely occluded.}
    \label{fig:rcnn}
\end{figure}
\section{Experiments} \label{sec:experiments}
%environment
We test the proposed method on a modified version of the \texttt{FetchPush-v0} environment from the OpenAI Gym using a Baxter robot. The goal of the agent is to push a object from its initial position to a goal position using its gripper.  Both the object and the target are randomly initialized within a rectangular region. The agent controls motion of the gripper along a  horizontal plane to the ground plane within a certain rectangular region, the gripper actions are thus $dx, dy$ of the gripper displacement. An inverse kinematics solver is used to compute the required joint angles to achieve this displacement.

The agent can similarly  move the camera along the horizontal plane, within a rectangular region. The camera can move up to 6\,cm per time step and not more than 20\,cm from the origin. 
%The camera is also constrained so that it cannot move too close to the object or target, since that would make the task too easy. 
We experimented with camera rotation in addition to just panning, but found that it did not help and was not used.  
There are randomly sampled distractors between the object and the camera. Each scene contains anywhere from 0 to 3 distractors. Each distractor is placed randomly within a rectangular region centered at the halfway point between the initial object location and the initial camera position. Each distractor is either a rectangular box, a truncated ellipsoid, or a cylinder. The color of the distractors is randomized. The camera is always initialized in the same starting location. 
 
The agent receives a $-1$ reward for every step it has not achieved the goal. When the object is within a tolerance radius 2\,cm from the goal location, the episode ends with success.  In all our results, we report the fraction of successful completions in 100 episodes in 3 training runs starting from different random seeds. 

Our experiments aim to answer the following questions:
\begin{enumerate}
\item How do object occlusions affect reinforcement learning manipulation policies?
\item Can active vision help reinforcement learning of manipulation policies in the presence of occlusions?
\item Does object-centric attention help and to what extent? 
\item Does curriculum learning in progressively more cluttered environments help policy learning under occlusions?
\item Are auxiliary visibility rewards necessary for learning active vision policies?
  
\end{enumerate}

We  compare the following architectures:
\begin{enumerate}
\item \textit{cam-static}: an agent that can only control its gripper, not its camera. Its architecture is depicted in Figure \ref{fig:arch} (d),(f). 
\item \textit{cam-static-image}: an agent that can only control its gripper, and its actor does not have an object detector module. Its  architecture is depicted in Figure \ref{fig:arch} (e), (f).
\item  \textit{cam-active-abstr}: an agent with hand/eye control and state abstraction.   Its  architecture is depicted in Figure \ref{fig:arch} (b), (c).
\item \textit{cam-active-full}: an agent with hand/eye control and without state abstraction. Its architecture is depicted in Figure \ref{fig:arch} (a), (c).
\item \textit{cam-random}: an agent with control over its gripper, with the camera action selected uniformly over the action space. It's architecture is depicted in Figure \ref{fig:arch} (d), (c).
\end{enumerate}

\begin{figure}[h]
\centering
\setlength{\tabcolsep}{0pt}
\begin{tabular}{cc}
  \includegraphics[width=0.5\textwidth]{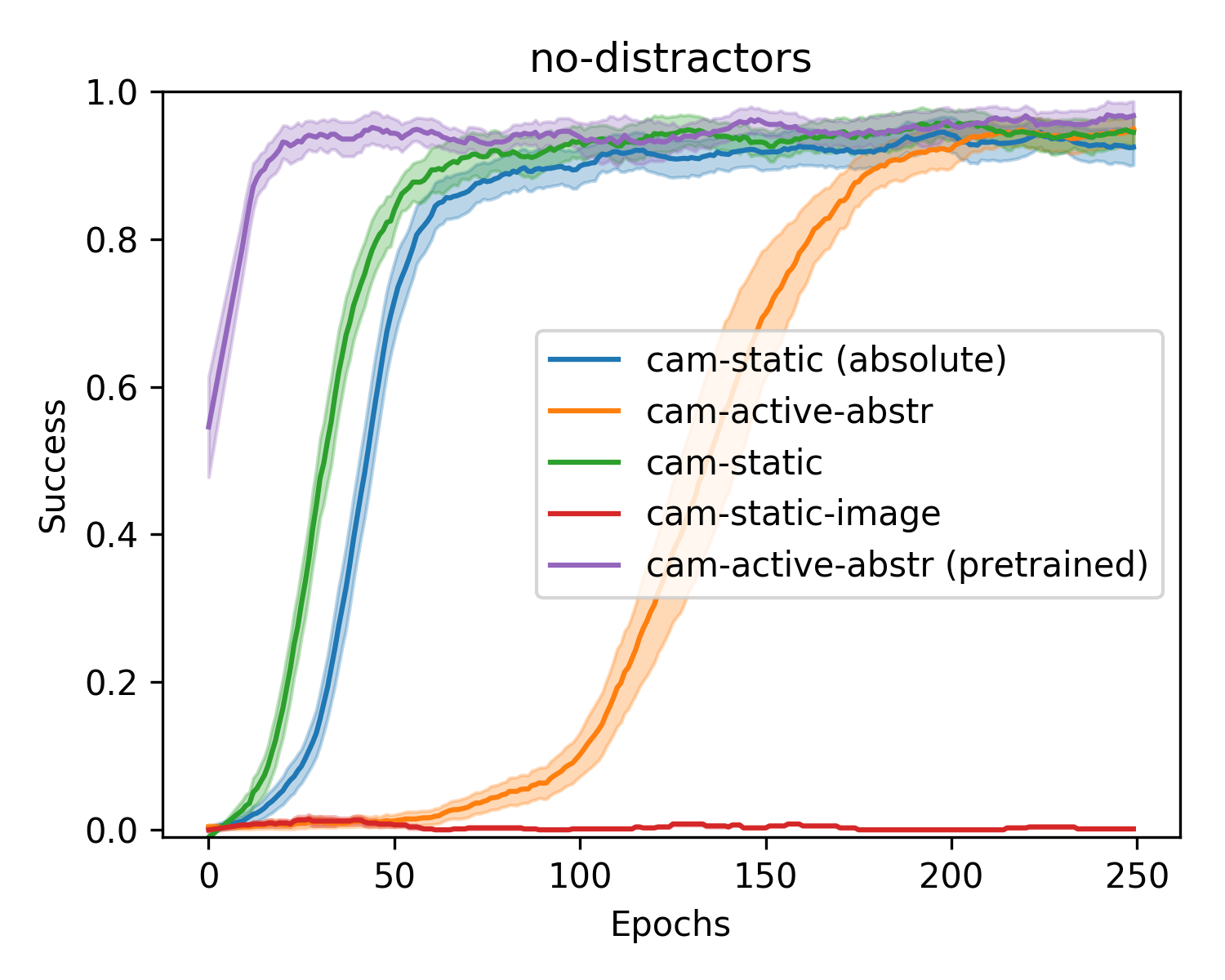} &  
  \includegraphics[width=0.5\textwidth]{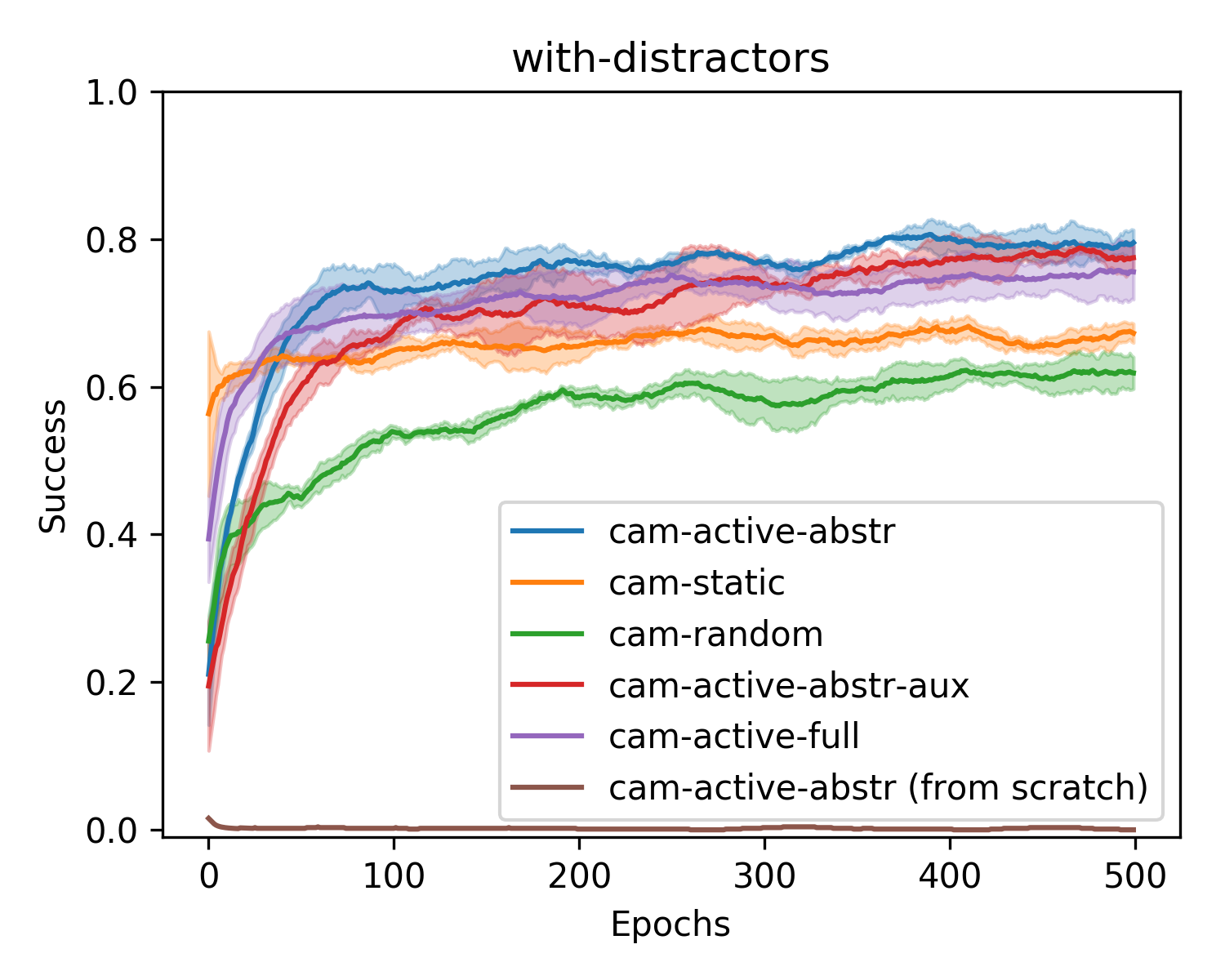} \\
\end{tabular}
\caption{\textit{Left:} \textbf{Environments without distractors.} Hand-eye policies train slower (\textit{cam-active-abstr}), yet all architectures achieve good asymptotic performance. Hand-eye policies can be effectively pretrained from hand only policies (\textit{cam-active-abstr} (pretrained)). Object-centric state encoding is beneficial (\textit{cam-static} outperforms \textit{cam-static (absolute)}). Finally, ignoring the location of the object of interest provided by the detector, and rather using only frame-centric appearance encoding does not result in successful behaviour (\textit{cam-static-image}). 
    \textit{Right:} \textbf{Environments with distractors.}
    Active vision helps to handle occlusions from distractors (\textit{cam-active-abstr} outperforms \textit{cam-static}). State abstraction helps for the hand actor policy (\textit{cam-active-abstr} outperforms \textit{cam-active-full}). Training directly in the environment with distractors, without pretraining on the easier environment does not result in successful behaviours (\textit{cam-active-abstr} (from scratch)). Auxiliary visibility reward is not helpful (\textit{cam-active-abstr-aux})necessary. A learned camera policy is superior to a random camera policy (\textit{cam-random}).
    Shaded area shows 1 standard error on the mean fraction of episodes which ended with success during training. We took the mean and computed the error over 20 episodes in each of 5 training runs using different seeds.
}
\label{fig:allresults}
\end{figure}

%\subsection{Training procedure}
\paragraph{Curriculum learning}
Interestingly, \textbf{we were unable to learn a successful manipulation policy in the environments with distractor objects} under any architecture, starting from random weights ---the detector submodule is excluded as it is always pretrained--- as shown In Figure \ref{fig:allresults}, right with the curve \textit{cam-active-abstr (from scratch)}. When the actor-critic network weights are pretrained using RL on environments without distractors, then many of our architectural variants were able to learn successful policies in the environments with distractors, as shown in Figure \ref{fig:allresults}.
Therefore, all our experiments in environments with distractors use pretraining  in environments without distractors.
As see in Figure \ref{fig:allresults} left, an agent with an active camera (\textit{cam-active-abstr}) is slower to train than an agent with a static camera (\textit{cam-static}). 
For training active camera policies, we found that training was greatly accelerated by first pretraining with a static camera, which means, ignoring the predicted camera actions, this is illustrated in Figure \ref{fig:allresults}, left, with the comparison between \textit{cam-active-abstr} and \textit{cam-active-abstr (pretrained)} curves. 

\paragraph{Object-centric representation}
Finally, we compare object-centric encoding of the state where the gripper and target 3D locations are provided relative to the 3D location of object of interest (Eq. \ref{eq:oc}) against an absolute state representation where gripper location, target location and object location are provided in absolute task space coordinates (Eq. \ref{eq:fc}). This comparison is illustrated in Figure \ref{fig:allresults}, left, with the comparison between \textit{cam-static (absolute)} and \textit{cam-static} curves. Since we find that the object-centric representation performs better, we use it for all subsequent experiments.

\begin{figure}[h]
\centering
\setlength{\tabcolsep}{0pt}
\begin{tabular}{ccccc}
 \includegraphics[width=0.2\textwidth]{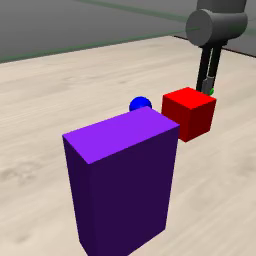} &  \includegraphics[width=0.2\textwidth]{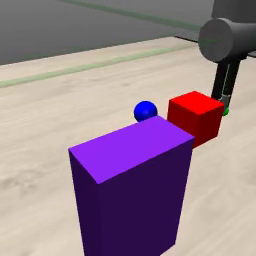} & 
 \includegraphics[width=0.2\textwidth]{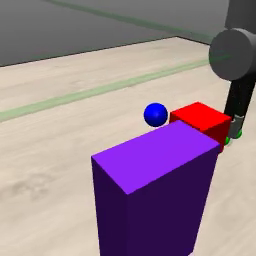} & 
 \includegraphics[width=0.2\textwidth]{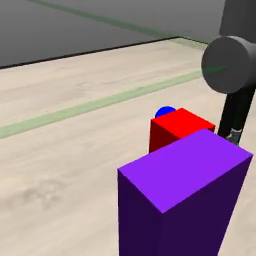} & 
 \includegraphics[width=0.2\textwidth]{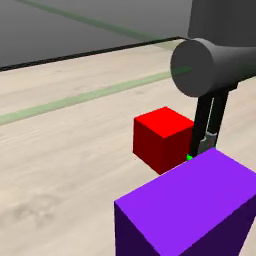} \\
 $\circ$ & $\leftarrow$ & $\leftarrow$ & $\nwarrow$ & $\uparrow$ \\
 \includegraphics[width=0.2\textwidth]{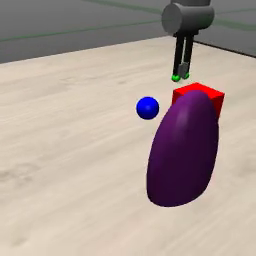} &  \includegraphics[width=0.2\textwidth]{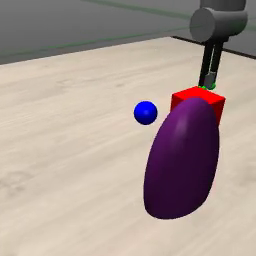} & 
 \includegraphics[width=0.2\textwidth]{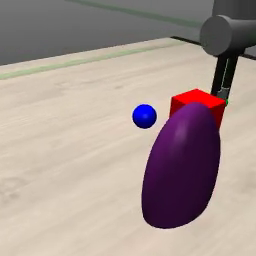} & 
 \includegraphics[width=0.2\textwidth]{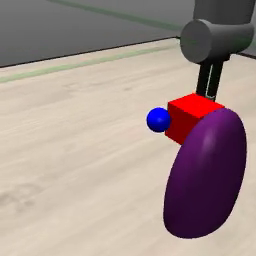} & 
 \includegraphics[width=0.2\textwidth]{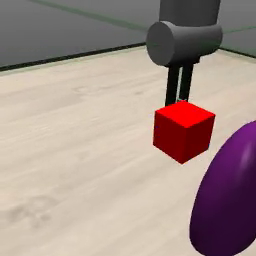} \\
 $\circ$ & $\circ$ & $\circ$ & $\leftarrow$ & $\nwarrow$ \\
 \includegraphics[width=0.2\textwidth]{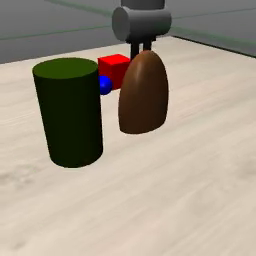} & 
 \includegraphics[width=0.2\textwidth]{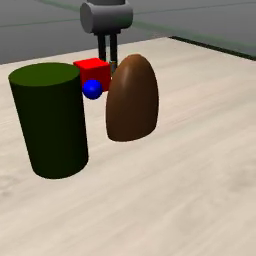} & 
 \includegraphics[width=0.2\textwidth]{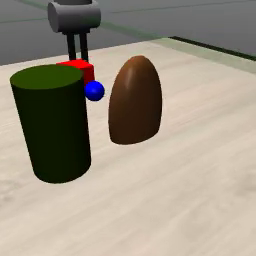} & 
 \includegraphics[width=0.2\textwidth]{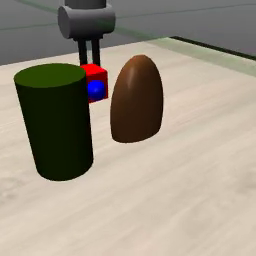} & 
 \includegraphics[width=0.2\textwidth]{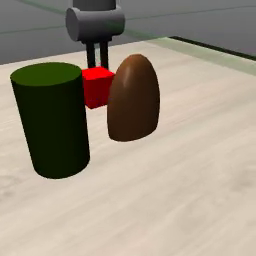} \\
 $\uparrow$ & $\rightarrow$ & $\rightarrow$ & $\rightarrow$ & $\circ$
\end{tabular}
\caption{ \textbf{Learned hand-eye control policies.} Each row corresponds to one episode, we show every other step of the episode. Since it is hard to tell the direction of camera movement from still frames, we draw arrows beneath each image showing the approximate direction of camera movement. In the top row, we see that the camera moves left and upwards to look over the obstacle. In the middle row, the robotic arm pushes the cube so that the left half is visible, and the camera moves left in order to expose the entire cube. In the bottom row, the cube is initially pushed leftward, so that if the camera is still, the cube would end up occluded by the cylinder. However, the camera moves right to compensate, so that the cube remains visible throughout the entire episode. }
\label{fig:video}
\end{figure}

\paragraph{Active vision for  manipulation  under  occlusions}
Though both active and static camera agents show similar success rates in the environments without distractors, (Figure \ref{fig:allresults} left),  agents that can control their camera   outperform agents with a static camera in environments with distractors, as shown in Figure \ref{fig:allresults}, right, with the comparison between \textit{cam-static} and \textit{cam-active-abstr} curves. The actor network that does not receive the frame appearance embedding learns faster than the variant that uses the frame appearance into the actor network. The fact that the \textit{cam-random} agent does not perform as well demonstrates that our \textit{cam-active-abstr} policy is learning something useful, and that simply having more viewpoints does not help in itself. 
We visualize learned active vision policies in Figure \ref{fig:video}. Video examples of the learned hand-eye coordination policies are available at \url{https://github.com/ricsonc/ActiveVisionManipulation}.

\paragraph{Auxiliary visibility reward}
We want to investigate whether auxiliary visibility rewards help to train active vision policies for manipulation. We compare the performance of an active camera controller trained under a combination of  manipulation reward and an auxiliary visibility reward being $0.25$  in value for every step that the  object detector detects the object successfully, against an active camera controller trained solely from the manipulation reward. We  expect the visibility reward to help the camera learn to move so that it increases visibility of the object of interest. In addition, the gripper may also aid the camera by pushing the object in a way which maximizes visibility. However, there is a potential downside that agent may favor actions less likely to push the object towards the target but rather improve visibility. We found such visibility reward did not help, as shown in Figure \ref{fig:allresults},right in the comparison between \textit{cam-active-abstr} and \textit{cam-active-abstr (visibility reward)}, as well as in the results of Table \ref{tab:performance}. This agrees with the observation from \cite{andrychowicz2017hindsight} that HER does not work well with dense rewards. 

\begin{table}[h!]
\centering
\begin{tabular}{c|c|c|c}
camera & abstr & vis reward & success \\ \hline
active & yes & no & $71.0\% \pm 2.2\%$ \\
active & yes & yes & $68.4\% \pm 1.5\%$ \\
active & no & no & $64.6\% \pm 1.5\%$ \\
random & yes & no & $55.7\% \pm 5.7\%$ \\
static & -- & no & $55.8\% \pm 6.6\%$ \\
\end{tabular}
\caption{ Success rate of the final policies at test time, averaged across 100 episodes from each of 5 training runs using different random seeds in environments with distractors. Hand-eye control policies with state abstraction for the gripper actor and no visibility auxiliary reward perform best. %\todo{ricson to write how success is measnured}
}
\label{tab:performance}
\end{table}

\section{Discussion - Future Work}
We proposed architectures for joint hand-eye coordination in the presence of environmental occlusions. We showed  active camera control policies can effectively anticipate occlusions due to hand movement and move accordingly to aid state estimation. This work attempts to stimulate research interest towards vision for action, as opposed to vision for recognition. To the best of our knowledge, this is the first work that jointly considers the learning problems of active perception and manipulation. 

Limitations of the current framework is that state estimation is memoryless, in other words, our object detectors operate independently at every frame. Future work includes combining active perception with visual memory architectures that encode a rich visual state that persist across timesteps of the episode,  and  integrates information of the visual scene in time. Another important and interesting direction is porting hand/eye coordination policies on a real robot for real time automated (un-instrumented) state estimation.

% \input{main.bbl}
%\bibliography{egbibnew,egbib_active} 
\bibliography{main.bib}
\clearpage

\end{document}